# On Information Regularization

**Adrian Corduneanu**
Artificial Intelligence Laboratory[1]
Massachussetts Institute of Technology
Cambridge, MA 02139

**Tommi Jaakkola**
Artificial Intelligence Laboratory
Massachussetts Institute of Technology
Cambridge, MA 02139

## Abstract

We formulate a principle for classification with the knowledge of the marginal distribution over the data points (unlabeled data). The principle is cast in terms of Tikhonov style regularization where the regularization penalty articulates the way in which the marginal density should constrain otherwise unrestricted conditional distributions. Specifically, the regularization penalty penalizes any information introduced between the examples and labels beyond what is provided by the available labeled examples. The work extends (Szummer and Jaakkola, 2003) to multiple dimensions, providing a regularizer independent of the covering of the space used in the derivation. In addition we lay the learning theoretical framework for classification with information regularization and provide a sample complexity bound. We illustrate the regularization principle in practice by restricting the class of conditional distributions to be logistic regression models and constructing the regularization penalty from a finite set of unlabeled examples.

## 1 INTRODUCTION

Consider the task of training a classifier from samples drawn according to a density $p(\mathbf{x}, y)$ over the joint space of data and class labels $\mathcal{X} \times \mathcal{Y}$. The task distinguishes itself from standard supervised learning in that additional abundant unlabeled data provides full knowledge of the marginal density $p(\mathbf{x})$ [2]. We investigate a principle for integrating this unlabeled information with minimal assumptions about the underlying density as introduced in (Szummer and Jaakkola, 2003), and we derive a regularizer of the conditional log-likelihood which complies to this principle.

The regularizer extends (Szummer and Jaakkola, 2003) in that it applies to any dimensionality and it is transparent to the covering of the space used in its derivation, while in one dimension it is analytically tractable. We show how the regularizer can be used to learn a classifier with no parametric assumptions about the conditional, but also provide practical algorithms when a parametric decision boundary is desirable. In the case of logistic regression we demonstrate that the unlabeled information can achieve a significant reduction in classification error. Finally, we provide a learning theoretical framework for learning a classifier under the presence of unlabeled information.

## 2 INFORMATION REGULARIZATION

The key question here is how to establish a general link between the marginal $p(\mathbf{x})$ and conditionals $p(y|\mathbf{x})$. A common direction in this regard tries to place the decision boundary, and therefore large changes in $p(y|\mathbf{x})$, in low density regions (Figure 1). In other words, tight clusters of points are likely to be labeled similarly, whereas the label may change across such clusters. It is less immediate how this intuition should translate into a formal relation between the marginal and the conditional.

We establish the relation by regularizing information. The key guiding principle here is that only labeled points can provide useful information about the conditional. By specifying any conditional function, i.e., $p(y|\mathbf{x})$ as a function of $\mathbf{x}$, we automatically introduce some information between the labels and examples. Such information, when not from the labeled examples, is artefactual and should be minimized. The regularization penalty should therefore be expressed in terms of mutual information.

In order to incorporate any known topological structure over the example space, however, we have to measure information locally. In other words, for any small region $Q$ defined in terms of the available metric, the regularization penalty should scale with $I_Q(y; \mathbf{x})$, the mutual information between $y$ and $\mathbf{x}$ restricted to region $Q$, where the marginal defined by $p(\mathbf{x})/p(Q)$. Moreover, the regulariza-tion penalty should scale with the probability mass per region, or, equivalently, be defined per point rather than per region (regions are secondary concepts). The local regularization penalty arising from the information regularization principle is therefore $p(Q)I_Q(y; \mathbf{x})$ over any (small) region $Q$.

---

[1] Work done while at University of Toronto
[2] We also consider the relaxation to finite-sample noisy estimates of the marginal

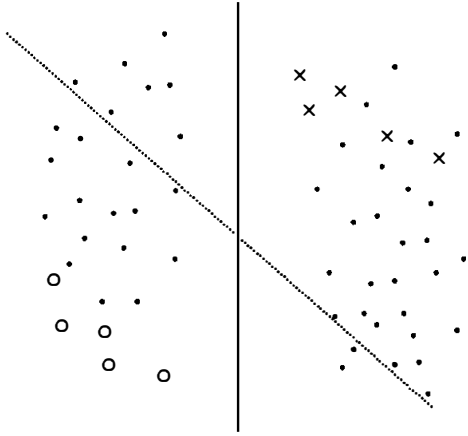

Figure 1: Principle Behind Information Regularization: Decision Boundaries Should not Cross Regions of High Data Density

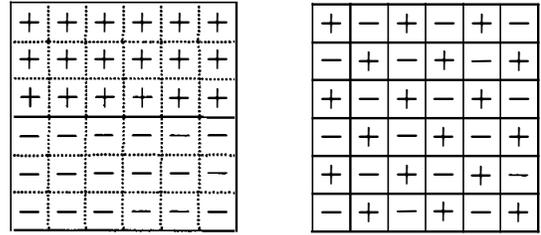

Figure 2: Complexity of Decision Boundaries not Captured by Mutual Information Between Data and Labels, Invariant to Permutations of Conditionals of Small Equally Probable Regions

## 2.1 LOCAL INFORMATION REGULARIZATION

In the absence of complete samples, unlabeled data provides no information about the conditional. Thus we would like to constrain the information $y$ provides about $\mathbf{x}$ in regions with no labeled samples. The relevant information theoretical quantity is

$$\log \frac{p_Q(\mathbf{x}, y)}{p_Q(\mathbf{x}) p_Q(y)}$$

where the subscript indicates restriction to the region $Q$. Its expected value over the region, the average mutual information $I_Q(\mathbf{x}, y)$, unfortunately is not indicative about *local* variations in conditional. Mutual information is invariant to permutations of conditionals of small regions of equal probability, thus constraining it does not enforce smoothness at the local level (Figure 2).

To derive an information-based measure of smoothness of $p(y|\mathbf{x})$ at the local level, we consider mutual information as the diameter of $Q$ approaches 0. If $\mathbf{x}_0$ is the expected value of $\mathbf{x}$ in the region, mutual information takes the following asymptotic form (see Appendix A for a derivation):

$$I_Q(\mathbf{x}; y) = \frac{1}{2} \text{Tr} \left[ \text{cov}_Q(\mathbf{x}) F(\mathbf{x}_0) \right] + \mathcal{O}\left( \text{diam}(Q)^3 \right) \quad (1)$$

where $F(\mathbf{x}) = \mathbf{E}_{p(y|\mathbf{x})} \left[ \nabla_\mathbf{x} \log p(y|\mathbf{x}) \cdot \nabla_\mathbf{x} \log p(y|\mathbf{x})^T \right]$ is the Fisher information and $\text{cov}_Q(\mathbf{x})$ is the covariance of $p_Q(\mathbf{x})$. In this derivation we have made the implicit assumption of differentiability of $p(y|\mathbf{x})$. This is not a practical restriction as functions with unbounded derivative can approximate a wide range of densities. Bounded variation will be a consequence of regularization rather than an assumption.

Note that since the covariance is $\mathcal{O}\left(\text{diam}(Q)^2\right)$, $I_Q(\mathbf{x}; y) \to 0$ as $\text{diam}(Q) \to 0$. As a measure of smoothness mutual information is not scale invariant, because there is not much uncertainty in $\mathbf{x}$ in an infinitesimal $Q$ to begin with. Therefore we normalize mutual information to $\text{diam}(Q)^2$ to characterize smoothness at local level. The actual normalization factor will not matter in the limit up to a multiplicative factor as long as it is $const \cdot \mathcal{O}\left(\text{diam}(Q)^2\right)$. Thus we may view it as normalization with respect to the variance of $\mathbf{x}$ while preserving the shape of $Q$ as $\text{diam}(Q) \to 0$, or *mutual information per unit variance*.

Finally, according to the stated principle we want to penalize more conditional changes in regions of high data density. This leads to the following regularizer in an infinitesimal region $Q$:

$$\frac{p(Q)}{\text{diam}(Q)^2} I_Q(\mathbf{x}; y) \approx \frac{1}{2} p(Q) \text{Tr} \left[ \frac{\text{cov}_Q(\mathbf{x})}{\text{diam}(Q)^2} F(\mathbf{x}_0) \right] \quad (2)$$

Let $Q_0$ be the shape similar to $Q$ such that $\text{diam}(Q_0) = 1$. As $Q$ shrinks $\mathbf{x}$ becomes uniform on $Q$ up to first order. Thus $\text{cov}_Q(\mathbf{x}) \approx \text{diam}(Q)^2 \text{cov}_{Q_0}$, where $\text{cov}_{Q_0}$ is the covariance of an uniform distribution on $Q_0$. The local regularizer becomes

$$\frac{1}{2} p(Q) \text{Tr} \left[ \text{cov}_{Q_0} F(\mathbf{x}_0) \right] \quad (3)$$

### 2.1.1 Shape-Independent Local Regularization

As introduced above local regularization depends on the shape of $Q$ through $\text{cov}_{Q_0}$. Symmetric regions like the sphere or the axis-parallel cube make this parameter a mul-tiple of identity, but the question is why should we prefer them. We introduce another principle that will remove this degree of freedom. Briefly, the regularizer must not a priori prefer a specific direction independent of $p(\mathbf{x})$ for the variation of the conditional. Formally, consider a small region $Q$ in which $p_Q(\mathbf{x})$ is uniform and $p(y=1|\mathbf{x}) = \mathbf{v} \cdot \mathbf{x} + c$ is linear, where $\mathbf{v}$ is the direction of highest variation. In this setting we have the following result:

**Theorem 1** *The local information regularizer is independent of $\mathbf{v}/\|\mathbf{v}\|$ if and only if $\text{cov}_{Q_0}$ is a multiple of the identity.*

**Proof** We have $F(\mathbf{x}_0) = \mathbf{v}\mathbf{v}^T$. The relevant quantity that should be independent of $\mathbf{v}/\|\mathbf{v}\|$ is therefore $\mathbf{v}^T \text{cov}_{Q_0} \mathbf{v}$. Let $v = \Lambda_i/\|\Lambda_i\|$, where $\Lambda_i$ is an eigenvector of $\text{cov}_{Q_0}$ of eigenvalue $\lambda_i$. Then $\mathbf{v}^T \text{cov}_{Q_0} \mathbf{v} = \lambda_i$ should not depend on the eigenvector. If follows that $\text{cov}_{Q_0}$ has equal eigenvalues, thus $\text{cov}_{Q_0} = \lambda I$. The converse is trivial. □

Dropping multiplicative constants, we have derived the following information regularizer on an infinitesimal region $Q$, where $\mathbf{x}_0$ is its center of mass:

$$p(Q)\text{Tr}\,[F(\mathbf{x}_0)] \tag{4}$$

## 2.2 GLOBAL INFORMATION REGULARIZATION

We derive a global regularizer of the log-likelihood that constrains the information $y$ provides about $\mathbf{x}$ and biases variations in the conditional to regions of low data density. The idea is to define a rich covering $\mathcal{X} = \cup_{Q \in \mathcal{Q}} Q$ with infinitesimal regions and sum the local regularizers over each region. The covering must satisfy certain properties, such as connectedness and a significant overlap between neighbors. This is because $p(y|\mathbf{x}_1)$ imposes a constraint on $p(y|\mathbf{x}_2)$ only through the regularizer, and only if $\mathbf{x}_1$ and $\mathbf{x}_2$ are in the same region, or are connected by a path of overlapping regions. Ideally, as the the size of the regions approaches 0 the overlap of neighbors approaches 100%. Note that with such overlap each point will belong to infinitely many regions, thus the sum of local regularizers will be infinity. We avoid over-counting by adjusting the weight of local regularizers.

In what follows we derive the regularizer from a specific covering; nevertheless, the limiting result will be the same for other coverings that abide to the above assumptions. $\mathcal{Q}$ consists of all axis-parallel cubes of length $l$ centered at the axis-parallel lattice points that are spaced at distance $l'$, where $l'$ is much smaller than $l$. As $l \to 0$ we would like the overlap factor $l/l'$ to approach infinity; for instance, $l' = l^2$. Each point belongs to $\lfloor l/l' \rfloor^d$ regions, where $d$ is the dimension of data, and this will be our discount factor to account for overlapping. Let $\mathcal{Q}'$ be the *partitioning* of $\mathcal{X}$ into atomic lattice cubes of length $l'$. Each region in $\mathcal{Q}$ is partitioned into $\lfloor l/l' \rfloor^d$ atomic cubes of $\mathcal{Q}'$, and each atomic cube is contained in $\lfloor l/l' \rfloor^d$ overlapping regions of $\mathcal{Q}$. We may now rewrite the global regularizer as a sum over the partition $\mathcal{Q}'$:

$$\begin{aligned}
\lim_{l \to 0} \sum_{Q \in \mathcal{Q}} \frac{p(Q)}{\lfloor l/l' \rfloor^d} \text{Tr}\,[F(\mathbf{x}_0(Q))] &= \\
\lim_{l \to 0} \sum_{Q' \in \mathcal{Q}'} p(Q') \sum_{Q \ni Q'} \frac{\text{Tr}\,[F(\mathbf{x}_0(Q))]}{\lfloor l/l' \rfloor^d} &= \\
\lim_{l' \to 0} \sum_{Q' \in \mathcal{Q}'} p(Q') \text{Tr}\,[F(\mathbf{x}_0(Q'))] &= \\
\int_{\mathcal{X}} p(\mathbf{x}) \text{Tr}\,[F(\mathbf{x})]\,d\mathbf{x}
\end{aligned} \tag{5}$$

Given labeled training data we can estimate the conditional by applying the information regularizer to the conditional log-likelihood:

$$\max_{\{p(y|\mathbf{x})\}} \sum \log p(y_i|\mathbf{x}_i) - \lambda \int_{\mathcal{X}} p(\mathbf{x}) \text{Tr}\,[F(\mathbf{x})]\,d\mathbf{x} \tag{6}$$

The maximum is over all continuous piecewise-differentiable conditionals subject to $0 \le p(y|\mathbf{x}) \le 1$ and $\sum_{y \in \mathcal{Y}} p(y|\mathbf{x}) = 1$. Full continuity is not necessary, but a continuous approximation to the discontinuity will always achieve a higher score. Note that on a continuous domain we cannot learn the conditional without the regularizer no matter how many labeled samples, because we make no other assumption about how conditionals at different locations relate to each other.

The positive $\lambda$ absorbs all constant multiplicative factors in the derivation, and also controls the strength of the regularization. At $\lambda \to \infty$ the penalty for any information in $y$ about $\mathbf{x}$ is high, and the maximizing conditional is the same at every location; its actual value depends on the overall number of training labeled samples in each class. At $\lambda \to 0$, we estimate each $p(y_i|\mathbf{x}_i)$ independently of unlabeled information (1 for continuous $\mathcal{X}$; the fraction of samples in class $y_i$ at $\mathbf{x}_i$ for discrete $\mathcal{X}$), then complete the conditional between training samples as if $p(\mathbf{x})$ is uniform. Only intermediate values of $\lambda$ make the variation of $p(y|\mathbf{x})$ depend on $p(\mathbf{x})$.

## 3 OPTIMIZATION WITH INFORMATION REGULARIZATION

We discuss several methods of optimizing the regularized likelihood (6) for continuous binary classification ($\mathcal{Y} = \{-1, 1\}$, continuous $\mathcal{X}$). To begin with we make no parametric assumptions about $p(y|\mathbf{x})$ and show that information regularization defines a unique solution. As in (Szummer and Jaakkola, 2003) we can use calculus of variations to obtain a differential equation that characterizes the optimal conditional. Given natural boundary conditions $p(\mathbf{x}) = 0$ and $\nabla_\mathbf{x} p(y|\mathbf{x}) = 0$ as well as the values of the conditional on all labeled samples, $p(y_i|\mathbf{x}_i) = p_0(y_i|\mathbf{x}_i)$, the conditional that minimizes the regularizer $\int p(\mathbf{x})\text{Tr}\,[F(\mathbf{x})]$ is a differential function (except maybe at the labeled samples, where it is only continuous) that satisfies the Euler-Lagrange condition:

$$\nabla_\mathbf{x} \log p(\mathbf{x}) \nabla_\mathbf{x} p(1|\mathbf{x})^T + \text{Tr}\,[\nabla^2_{\mathbf{xx}} p(1|\mathbf{x})] + \frac{1}{2} \frac{p(1|\mathbf{x}) - p(-1|\mathbf{x})}{p(1|\mathbf{x})p(-1|\mathbf{x})} \|\nabla_\mathbf{x} p(1|\mathbf{x})\|^2 = 0 \quad (7)$$

This equation uses the unlabeled information $\nabla_\mathbf{x} \log p(\mathbf{x})$ to complete the conditional from its value on labeled samples in a unique way. If $I(\{p_0(y_i|\mathbf{x}_i)\})$ is the minimal regularizer given the value of $p(y|\mathbf{x})$ on the labeled samples, to optimize (6) we need to consider all such values:

$$\max_{\{p_0(y_i|\mathbf{x}_i)\}\in[0,1]^n} \sum \log p_0(y_i|\mathbf{x}_i) - \lambda I(\{p_0(y_i|\mathbf{x}_i)\}) \quad (8)$$

In one dimension the differential equation (7) can be solved analytically. If $f(x) = p(1|x)$, rewrite the equation as

$$\frac{d}{dx} \log \left[ p(x) \frac{|\frac{df}{dx}|}{\sqrt{f(1-f)}} \right] = 0 \quad (9)$$

thus $\frac{1}{\sqrt{f(1-f)}}|\frac{df}{dx}| = \frac{c}{p(x)}, c \geq 0$. Therefore on each interval with non-zero $\frac{df}{dx}$ we have $\frac{1}{\sqrt{f(1-f)}}\frac{df}{dx} = \frac{c}{p(x)}$ for some (positive or negative) $c$. The left hand side can be integrated analytically to $-2\arctan\sqrt{\frac{1}{f} - 1}$, therefore:

$$p(y = 1|x) = \frac{1}{1 + \tan^2\left(-c \int \frac{1}{p(x)}\right)} \quad (10)$$

where $c$ and the additive constant in $\int 1/p$ can be determined from the values of the conditional at labeled samples. $c$ can change at the points where $\frac{df}{dx} = 0$. The conditional can be computed exactly provided $1/p$ is analytically integrable, or approximated numerically from $\int_a^x 1/p$.

In Figure 3 we show the effect of various data densities on the solution. Note that if $p(x)$ is uniform the conditional is close to but not linear, as the variation is penalized more when $p(1|x)$ is close to 0 or 1 rather then around 0.5.

Solving (7) and (8) numerically in high dimensions is a complex task, and we need simplifying assumptions about $p(y|\mathbf{x})$ and $p(\mathbf{x})$ for tractable optimization. We consider parametric representations of the decision boundary, as well as kernel estimates of $p(\mathbf{x})$ from a finite unlabeled sample.

### 3.1 PARAMETRIC DECISION BOUNDARY

One of the merits of information regularization is that no parametric model is necessary to propagate unlabeled information. Unlabeled data and few labeled samples provide a decision boundary with minimal assumptions about $p(y|\mathbf{x})$ or how it relates to $p(\mathbf{x})$. Nevertheless, we can still employ information regularization on discriminative tasks in which a parametric decision boundary is desired. The goal is to estimate the conditional $p(y|\mathbf{x}; \theta)$ from a family parametrized by $\theta$ given unlabeled information $p(\mathbf{x})$. We can apply the information regularizer as before, but optimizing over $\theta$:

$$\max_\theta \sum \log p(y_i|\mathbf{x}_i; \theta) - \lambda \int_\mathcal{X} p(\mathbf{x})\text{Tr}\,[F(\mathbf{x}; \theta)]\,d\mathbf{x} \quad (11)$$

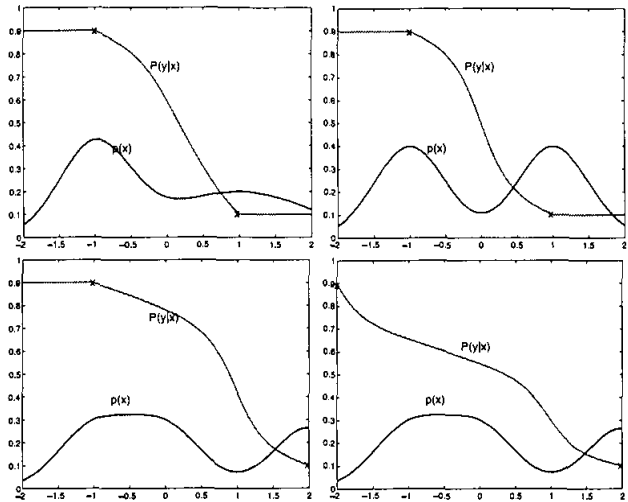

Figure 3: Conditionals that Minimize the Information Regularizer for Various One-Dimensional Data Densities While the Value at Boundary Labeled Points Is Fixed

We illustrate this approach on logistic regression, in which we restrict the conditional to linear decision boundaries with the following parametric form: $p(y|\mathbf{x}; \theta) = \sigma(y\theta^T\mathbf{x})$, where $y \in \{-1, 1\}$ and $\sigma(x) = 1/(1 + \exp(-x))$. We get $F(\mathbf{x}; \theta) = \sigma(\theta^T\mathbf{x})\sigma(-\theta^T\mathbf{x})\theta\theta^T$ and the regularizer

$$\|\theta\|^2 \int p(\mathbf{x})\sigma(\theta^T\mathbf{x})\sigma(-\theta^T\mathbf{x})d\mathbf{x} \quad (12)$$

The term $\sigma(\theta^T\mathbf{x})\sigma(-\theta^T\mathbf{x}) = p(y|\mathbf{x})p(\tilde{y}|\mathbf{x})$ focuses on the decision boundary. Therefore compared to the standard logistic regression regularizer $\|\theta\|^2$, we penalize more decision boundaries crossing regions of high data density. Note that the term also makes the regularizer non-convex, making optimization potentially more difficult. This lack of convexity is however unavoidable by any algorithm using unlabeled information, that should take into account complex multi-modal data densities.

### 3.2 FINITE UNLABELED DATA APPROXIMATIONS

To finalize a practical formulation of the optimization we must provide an approximate regularizer from a large but finite unlabeled sample $\{\mathbf{x}'_j\}$ rather than full knowledge of $p(\mathbf{x})$. We consider the empirical approximation $\frac{1}{m}\sum \delta(\mathbf{x} - \mathbf{x}'_j)$, kernel density estimators, as well as parametric models.

The empirical approximation can only be used in finite domains or when the conditional is parametrized; otherwise regions of zero probability make the conditional arbitrary in (8) except on labeled samples. In logistic regression however, where all conditionals are tied through $\theta$, the counting approximation becomes relevant:

$$\max_{\boldsymbol{\theta}} \sum_i \log \sigma(y_i \boldsymbol{\theta} \mathbf{x}_i) - \frac{\lambda}{m} \sum_j \sigma(\boldsymbol{\theta}^T \mathbf{x}'_j) \sigma(-\boldsymbol{\theta}^T \mathbf{x}'_j) \quad (13)$$

This criterion can be easily optimized by gradient-ascent or Newton type algorithms. In the results section we also demonstrate optimization by continuation, in which $\lambda$ is gradually increased while following the solution.

If unlabeled data is limited, we may prefer a kernel estimate $p(\mathbf{x}) = \frac{1}{m}\sum_{j=1}^{m} K(\mathbf{x}, \mathbf{x}'_j)$ to the empirical approximation, provided the regularization integral remains tractable. In the regularization of logistic regression, if the kernels are Gaussian we can make the integral tractable by approximating $\sigma(\boldsymbol{\theta}^T \mathbf{x})\sigma(-\boldsymbol{\theta}^T \mathbf{x})$ with a degenerate Gaussian. Either from the Laplace approximation, or the Taylor expansion $\log(1 + e^x) \approx \log 2 + x/2 + x^2/8$, we derive the following approximation:

$$\sigma(\boldsymbol{\theta}^T \mathbf{x})\sigma(-\boldsymbol{\theta}^T \mathbf{x}) \approx \frac{1}{4}\exp\left(-\frac{1}{4}(\boldsymbol{\theta}^T \mathbf{x})^2\right) \quad (14)$$

With this approximation computing the integral of the regularizer over the kernel at $\mu$ of variance $\tau \mathbf{I}$ becomes integration of a Gaussian:

$$\frac{1}{4}\exp\left(-\frac{1}{4}(\boldsymbol{\theta}^T \mathbf{x})^2\right) \mathcal{N}(\mathbf{x}; \boldsymbol{\mu}, \tau \mathbf{I}) =$$

$$\frac{1}{4}\sqrt{\frac{\det \Sigma_\theta}{\det \tau \mathbf{I}}}\exp\left(-\frac{\boldsymbol{\mu}^T(\tau \mathbf{I} - \Sigma_\theta)\boldsymbol{\mu}}{2\tau^2}\right) \mathcal{N}\left(\mathbf{x}; \frac{\Sigma_\theta \boldsymbol{\mu}}{\tau}, \Sigma_\theta\right)$$

where
$$\Sigma_\theta = \left(\frac{1}{\tau}\mathbf{I} + \frac{1}{2}\boldsymbol{\theta}\boldsymbol{\theta}^T\right)^{-1} = \tau\left[\mathbf{I} - \frac{1}{2}\boldsymbol{\theta}\boldsymbol{\theta}^T / \left(\frac{1}{\tau} + \frac{1}{2}\|\boldsymbol{\theta}\|^2\right)\right]$$

After integration only the multiplicative factor remains:

$$\frac{1}{4}\left(1 + \frac{\tau}{2}\|\boldsymbol{\theta}\|^2\right)^{-\frac{1}{2}}\exp\left(-\frac{1}{4}\frac{(\boldsymbol{\theta}^T \boldsymbol{\mu})^2}{1 + \frac{\tau}{2}\|\boldsymbol{\theta}\|^2}\right)$$

Therefore if we place a Gaussian kernel of variance $\tau \mathbf{I}$ at each unlabeled sample $\mathbf{x}'_j$ we obtain the following approximation to (12):

$$\frac{\|\boldsymbol{\theta}\|^2}{\sqrt{1 + \frac{\tau}{2}\|\boldsymbol{\theta}\|^2}}\frac{1}{4m}\sum_{j=1}^{m}\exp\left(-\frac{1}{4}\frac{(\boldsymbol{\theta}^T \mathbf{x}'_j)^2}{1 + \frac{\tau}{2}\|\boldsymbol{\theta}\|^2}\right) \quad (15)$$

This regularizer can be also optimized by gradient ascent or Newton's method.

## 4 LOGISTIC REGRESSION EXPERIMENTS

We illustrate the application of information regularization to synthetic classification tasks. We generate data from two bivariate Gaussian densities of equal covariance, a model in which the linear decision boundary of logistic regression can be Bayes optimal. However, the small number of labeled samples is not enough to accurately estimate the model, and we show that information regularization with unlabeled data can significantly improve error rates.

We compare a few criteria: logistic regression trained only on labeled data and regularized with the standard $\|\boldsymbol{\theta}\|^2$; logistic regression regularized with the information regularizer derived from the empirical estimate to $p(\mathbf{x})$ (13) ; and logistic regression with the information regularizer derived from a Gaussian kernel estimate of $p(\mathbf{x})$ (15).

We have optimized the regularized likelihood $L(\boldsymbol{\theta})$ both with gradient ascent $\boldsymbol{\theta} \leftarrow \boldsymbol{\theta} + \alpha \nabla_\theta L(\boldsymbol{\theta})$, and with Newton's method (iterative re-weighted least squares) $\boldsymbol{\theta} \leftarrow \boldsymbol{\theta} - \alpha \nabla_{\theta\theta}^2 L(\boldsymbol{\theta})^{-1} \nabla_\theta L(\boldsymbol{\theta})$ with similar results. Newton's method converges with fewer iterations, but computing the Hessian becomes prohibitive if data dimensionality is high, and convergence depends on stronger assumptions that those for gradient ascent. Gradient ascent is safer, but slower if not too many parameters.

We ran multiple experiments (100) with data drawn from the same model and averaged the error rates to obtain statistically significant results. In Figure 4 we have obtained the error rates on 5 labeled and 100 unlabeled samples. On each data set we initialized the iteration randomly multiple times. The information regularizers derived from kernel and empirical estimates perform indistinguishable on such large number of unlabeled samples. They both outperform the standard labeled regularization significantly.

## 5 INFORMATION REGULARIZATION AND LEARNING THEORY

We provide a theoretical framework for learnability under information regularization, and asses the sample complexity of learning. While the learning framework is general, we derive sample-size bounds only for square loss and one-dimensional $\mathcal{X}$ and binary $\mathcal{Y}$, and discuss possible extensions.

To build a learning theory we need to formalize the learned concepts, the concept class (from which to learn them), and a measure of achievement consistent with (6). The key is then to show that the task is learnable in terms of the complexity of the concept class.

Standard PAC-learning of indicator functions of class membership will not suffice for our purpose. Indeed, con-ditionals with very small information regularizer can still have very complex decision boundaries, of infinite VC-dimension. Instead, we rely on the *p-concept* (Kearns and Schapire, 1994) model of learning full conditional densities: concepts are functions $h(y|\mathbf{x}) : \mathcal{X} \to [0,1]$. Then the concept class is that of conditionals with bounded information regularizer:

$$\mathcal{I}_\gamma(p) = \left\{ h : \int_\mathcal{X} p(\mathbf{x}) \sum_y h(y|\mathbf{x}) \|\nabla_\mathbf{x} \log h(y|\mathbf{x})\|^2 \leq \gamma \right\}$$

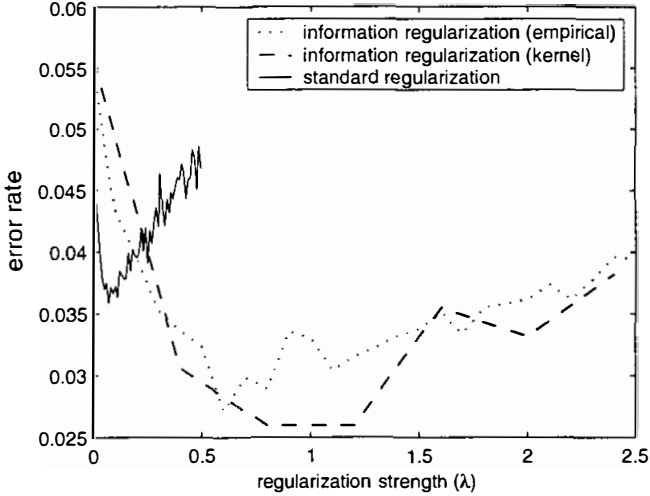

Figure 4: Average Error Rates of Logistic Regression with and without Information Regularization on 100 Random Selections of 5 Labeled and 100 Unlabeled Samples from Bivariate Gaussian Classes

We measure the quality of learning by a loss function $L_h : \mathcal{X} \times \mathcal{Y} \to [0, \infty)$. This can be the log-loss $-\log h(y|\mathbf{x})$ associated with maximizing likelihood, or the square loss $(h(y|\mathbf{x}) - 1)^2$ whose advantage is boundedness. The goal is to estimate from a labeled sample a concept $p_{opt}$ [3] from $\mathcal{I}_\gamma(p)$ that minimizes the expected loss $\mathrm{E}_{p(\mathbf{x})p(y|\mathbf{x})}[L_h]$.

One cannot compute the expected loss directly because it depends on the unknown $p(y|\mathbf{x})$. To optimize it, we minimize the empirical loss instead (log-likelihood if log-loss)

$$\hat{h} = \arg\min_h \hat{\mathrm{E}}[L_h] = \arg\min_h \frac{1}{n} \sum_{i=1}^n L_h(\mathbf{x}_i, y_i)$$

We say the task is learnable if with high probability in the sample the empirical loss converges to the true loss uniformly for all concepts as $n \to \infty$. This guarantees that $\mathrm{E}[L_{\hat{h}}]$ approximates $\mathrm{E}[L_{p_{opt}}]$ well. Formally,

$$\Pr[\exists h \in \mathcal{I}_\gamma(p) : |\hat{\mathrm{E}}[L_h] - \mathrm{E}[L_h]| > \epsilon] \leq \delta \quad (16)$$

where the probability is with respect to all samples of size $n$. The inequality should hold for $n$ polynomially large in $1/\epsilon, 1/\delta, 1/\gamma$.

---

[3] $p_{opt}(y|\mathbf{x})$ is $p(y|\mathbf{x})$ only when $p(y|\mathbf{x}) \in \mathcal{I}_\gamma(p)$

## 5.1 MEASURES OF COMPLEXITY

The sample size for a desired learning accuracy will be a function of the *complexity* of $\mathcal{I}_\gamma(p)$, like VC-dimension in PAC-learning. One such measure is the bound on the information regularizer $\gamma$; however, we should also take into account the complexity of $p(x)$.

Intuitively, learning is difficult when significant probability mass lies in regions of small $p(x)$ where the variation of $h$ is less constrained. Learning is also difficult when $p(x)$ has many modes of high probability separated by low probability, because the variation of $h$ is constrained only within each region. We define two quantities to characterize the complexity of $p(x)$. For each $\alpha \in [0, 1)$ let $M_p(\alpha) = \{x : p(x) \leq \alpha\}$ be the points of density below $\alpha$. Let $m_p(\alpha) = \Pr[M_p(\alpha)]$ be the total mass of small density. Let $C_p(\alpha)$ be the partition of $\mathcal{X} \setminus M_p(\alpha)$ into maximal disjoint intervals, and $c_p(\alpha)$ its count. We will provide a learning bound in terms of $m_p(\alpha)$, $c_p(\alpha)$, and $\gamma$.

The two measures of complexity are well-behaved for the useful densities. Densities of bounded support, Laplace and Gaussian, as well mixtures of these have $m_p(\alpha) < K\alpha$. Mixtures of single-mode densities have $c_p(\alpha)$ bounded by the number of mixtures.

## 5.2 DERIVATION OF A LEARNING BOUND

We derive the following sample complexity bound:

**Theorem 2** *Let $\epsilon, \delta > 0$. Then*

$$\Pr[\exists h \in \mathcal{I}_\gamma(p) : |\hat{\mathrm{E}}[L_h] - \mathrm{E}[L_h]| > \epsilon] < \delta$$

*where the probability is over samples of size $n$ greater than*

$$\mathcal{O}\left(\frac{1}{\epsilon^4}\left(\log\frac{1}{\epsilon}\right)\left[\log\frac{1}{\delta} + c_p(m_p^{-1}(\epsilon^2)) + \frac{\gamma}{(m_p^{-1}(\epsilon^2))^2}\right]\right)$$

Had $\mathcal{I}_\gamma(p)$ been finite, we would have derived a learning result from McDiarmid's inequality (McDiarmid, 1989) and the union bound as in (Haussler, 1990):

$$\Pr[\exists h \in \mathcal{I}_\gamma(p) : |\hat{\mathrm{E}}[L_h] - \mathrm{E}[L_h]| > \epsilon] \leq 2|\mathcal{I}_\gamma(p)|e^{-2\epsilon^2 n} \quad (17)$$

Hence the idea of replacing $\mathcal{I}_\gamma(p)$ with a finite discretization $\mathcal{I}_\gamma^\epsilon(p)$ for which the above inequality holds. If for any $h$ in $\mathcal{I}_\gamma(p)$ its representative $q_h$ from the discretization is guaranteed to be "close", and if $|\mathcal{I}_\gamma^\epsilon(p)|$ is small enough, we can extend the learning result from finite sets with

$$|\hat{\mathrm{E}}[L_h] - \mathrm{E}[L_h]| \leq |\hat{\mathrm{E}}[L_h] - \hat{\mathrm{E}}[L_{q_h}]| + \\ + |\mathrm{E}[L_h] - \mathrm{E}[L_{q_h}]| + |\hat{\mathrm{E}}[L_{q_h}] - \mathrm{E}[L_{q_h}]| \quad (18)$$

To discretize $\mathcal{I}_\gamma(p)$ we choose some $M$ points from $\mathcal{X}$ and discretize possible values of $h$ at those points into $1/\tau$ intervals of length $\tau > 0$. Any $h$ is then represented by one of $(1/\tau)^M$ combinations of small intervals. $\mathcal{I}_\gamma^\epsilon(p)$ consists of one representative from $\mathcal{I}_\gamma$ corresponding to each such combination (provided it exists). It remains to select the $M$ points and $\tau$ to guarantee that $h$ and $q_h$ are "close", and $|\mathcal{I}_\gamma^\epsilon(p)| = (1/\tau)^M$ is small.

Our starting point is Lemma 3 from Appendix B that bounds the variation of $h$ on an interval in terms of its information regularizer and $\int 1/p$. We can use it to bound $(h(1|x) - h'(1|x))^2$ on an interval $(x_1, x_2)$ independently of $h, h' \in \mathcal{I}_\gamma(p)$, provided $h, h'$ are within $\tau$ of each other at the endpoints, and $\int_{x_1}^{x_2} dx/p(x)$ is small. If we select the $M$ points of $\mathcal{I}_\gamma^\epsilon$ to make $\int 1/p$ small on each interval of the partition (except on the tail $M_p(\alpha)$), we can quantify the "closeness" of $h$ and $q_h$ as in Theorem 5:

$$|\mathrm{E}[L_h] - \mathrm{E}[L_{q_h}]| \le 2\left[m_p(\alpha) + \frac{\gamma}{2N^2\alpha^2} + 3\tau\right]^{1/2}$$

$$|\hat{\mathrm{E}}[L_h] - \hat{\mathrm{E}}[L_{q_h}]| \le 2\left[\bar{\epsilon} + m_p(\alpha) + \frac{\gamma + \gamma N\bar{\epsilon}}{2N^2\alpha^2} + 3\tau\right]^{1/2}$$

with probability at least $1 - (M+1)\exp(-2\bar{\epsilon}^2 n)$, where $\alpha \in (0, 1)$ is a free parameter to be optimized later, and $N = M + 1 - 2c_p(\alpha)$. We can combine the last two inequalities and (17) in (18) and optimize over $M, \tau, \alpha, \bar{\epsilon}$ to obtain a learning result.

To derive a general result (without knowing $m_p(\alpha), c_p(\alpha)$) we must choose possibly non-optimal values of the free parameters. If $N = \frac{\gamma}{2\alpha^2}$, $\bar{\epsilon} = \epsilon^2$, $\tau = \epsilon^2$, $m_p(\alpha) = \epsilon^2$, we obtain the asymptotic sample size stated in the theorem.

#### 5.2.1 Extensions

We consider extensions of the sample-complexity bound to multiclass classification, multidimensional $\mathcal{X}$, and log-loss (maximum likelihood) instead of square loss. To extend the results to the unbounded log-loss, we can use the equivalence results between square loss and log loss are presented in (Abe, Takeuchi, and Warmuth, 2001), where the $\epsilon$-Bayesian averaging trick effectively renders log-loss bounded. To extend the results to multiple dimensions we need a multidimensional equivalent of Lemma 3. Although intuitively feasible, such result could be difficult to obtain.

## 6 DISCUSSION

We have extended information regularization in several respects. We formulated the principle as a Tikhonov style regularization, providing a continuous version of the regularization penalty in multiple dimensions (independent of any topological cover used in a finite approximation). We also derived the differential equation governing the minimum penalty interpolation between the conditionals, where the interpolating solution can be found in closed form in one dimension. One way to reap the benefits of the new regularization principle in practice (without having to solve a multi-dimensional differential equation) is to formulate the regularization problem within a limited class of parameterized conditionals such as the logistic regression models we used here.

We showed that the regularization penalty serves as a valid notion of complexity of learning with unlabeled data, where the complexity measure depends both on the conditionals as well as the marginal distribution. Non-parametric tasks become learnable under no other assumptions but those imposed by the information regularizer.

## A  ASYMPTOTICS OF MUTUAL INFORMATION

We derive an asymptotic formula for the mutual information between data and labels in a region $Q$ as the scale $R$ (radius, diameter) approaches 0. We begin with the defining formula for mutual information:

$$I_Q(\mathbf{x}; y) = \sum_{y \in \mathcal{Y}} \int_{\mathbf{x} \in Q} p_Q(\mathbf{x}) p(y|\mathbf{x}) \log \frac{p(y|\mathbf{x})}{p_Q(y)} d\mathbf{x} \quad (19)$$

where the subscript $Q$ indicates that the joint is restricted to the region: $p_Q(\mathbf{x}) = p(\mathbf{x})/\int_{\mathbf{x} \in Q} p(\mathbf{x})d\mathbf{x}$, and $p_Q(y) = \int_{\mathbf{x} \in Q} p_Q(\mathbf{x}) p(y|\mathbf{x}) d\mathbf{x}$.

Let $\mathbf{x}_0 = \mathrm{E}_{p_Q(\mathbf{x})}[\mathbf{x}]$ be the average value of $\mathbf{x}$ in the region. To simplify notation let $G = \nabla_{\mathbf{x}} p(y|\mathbf{x}_0)$ and $H = \nabla^2_{\mathbf{xx}} p(y|\mathbf{x}_0)$ be the gradient and the Hessian of the conditional at $\mathbf{x}_0$. The conditional has the following second order Taylor expansion about $\mathbf{x}_0$:

$$p(y|\mathbf{x}) = p(y|\mathbf{x}_0) + G^T(\mathbf{x} - \mathbf{x}_0) + \\ + (\mathbf{x} - \mathbf{x}_0)^T H(\mathbf{x} - \mathbf{x}_0) + \mathcal{O}(R^3) \quad (20)$$

By taking expectation with respect to $p_Q(\mathbf{x})$ we also get $p_Q(y) = p(y|\mathbf{x}_0) + \mathrm{Tr}[\mathrm{cov}_Q(\mathbf{x})H] + \mathcal{O}(R^3)$ where $\mathrm{cov}_Q(\mathbf{x})$ is the covariance of $p_Q(\mathbf{x})$. Next we use $1/(1+x) = 1 - x + x^2 + \mathcal{O}(x^3)$ and $\log(1+x) = x - x^2/2 + \mathcal{O}(x^3)$ to obtain:

$$\log \frac{p(y|\mathbf{x})}{p_Q(y)} = \frac{1}{p(y|\mathbf{x}_0)} \big[G^T(\mathbf{x} - \mathbf{x}_0) + \\ + (\mathbf{x} - \mathbf{x}_0)^T H(\mathbf{x} - \mathbf{x}_0) - \mathrm{Tr}[\mathrm{cov}_Q(\mathbf{x})H] - \\ - [G^T(\mathbf{x} - \mathbf{x}_0)]^2/2p(y|\mathbf{x}_0)\big] + \mathcal{O}(R^3) \quad (21)$$

We only need to multiply the above equation by the expansion of $p(y|\mathbf{x})$ again and take the expectation with respect to $p_Q(\mathbf{x})$ to get:

$$I_Q(\mathbf{x}; y) = \sum_{y \in \mathcal{Y}} \frac{1}{2} p(y|\mathbf{x}_0) \mathrm{Tr}\left[\mathrm{cov}_Q(\mathbf{x}) G G^T\right] + \mathcal{O}(R^3) \quad (22)$$

Notice that $\sum_y p(y|\mathbf{x}_0) G G^T$ is just the Fisher information at $\mathbf{x}_0$. Finally

$$I_Q(\mathbf{x}; y) = \frac{1}{2} \mathrm{Tr}\left[\mathrm{cov}_Q(\mathbf{x}) F(\mathbf{x}_0)\right] + \mathcal{O}(R^3) \quad (23)$$

## B  TECHNICAL RESULTS FOR LEARNING THEORY

**Lemma 3** *For $x_1 < x_2$, $\mathcal{Y} = \{-1, 1\}$*

$$\int_{x_1}^{x_2} p(x) \mathrm{E}\left[\left(\frac{d}{dx}\log h(y|x)\right)^2\right] \geq \frac{4(h(1|x_1) - h(1|x_2))^2}{\int_{x_1}^{x_2}\frac{1}{p(x)}dx}$$

*where the expectation is with respect to $h(y|x)$.*

**Proof** After rewriting the expected value we use Cauchy-Schwartz, then $h(1|x)h(-1|x) \leq \frac{1}{4}$:

$$\int_{x_1}^{x_2}\frac{1}{p(x)}dx \cdot \int_{x_1}^{x_2} p(x)\frac{\left(\frac{d}{dx}h(1|x)\right)^2}{h(1|x)h(-1|x)}dx \geq$$

$$\left(\int_{x_1}^{x_2}\frac{\frac{d}{dx}h(1|x)}{\sqrt{h(1|x)h(-1|x)}}dx\right)^2 \geq 4\left(\int_{x_1}^{x_2}\frac{d}{dx}h(1|x)dx\right)^2$$

□

**Lemma 4** *The square loss $L_h = (h(y|x) - 1)^2$ satisfies*

$$|\mathrm{E}[L_{h_1}] - \mathrm{E}[L_{h_2}]| \leq 2\mathrm{E}\left[(h_1(1|x) - h_2(1|x))^2\right]^{\frac{1}{2}}$$

$$|\hat{\mathrm{E}}[L_{h_1}] - \hat{\mathrm{E}}[L_{h_2}]| \leq 2\left[\frac{1}{n}\sum_{i=1}^{n}(h_1(1|x_i) - h_2(1|x_i))^2\right]^{\frac{1}{2}}$$

**Proof** A simple application of Cauchy's inequality. □

**Theorem 5** *For every $\alpha \in (0,1)$ and $M$ there exist points $\{x_1, x_2, \ldots, x_M\}$ from $\mathcal{X}$ such that any $h_1, h_2 \in \mathcal{I}_\gamma(p)$ with $|h_1(1|x_i) - h_2(1|x_i)| \leq \tau, i = 1 \ldots M, \tau \in (0,1)$ satisfy*

$$|\mathrm{E}_{p(x)}[L_{h_1}] - \mathrm{E}_{p(x)}[L_{h_2}]| \leq 2\left[m_p(\alpha) + \frac{\gamma}{2N^2\alpha^2} + 3\tau\right]^{1/2}$$

*where $N = M + 1 - 2c_p(\alpha)$. Also, with probability at least $1 - (M+1)\exp(-2\epsilon^2 n)$ over a sample of size $n$ from $\mathcal{X}$, for any such $h_1$ and $h_2$ we have:*

$$|\hat{\mathrm{E}}[L_{h_1}] - \hat{\mathrm{E}}[L_{h_2}]| \leq 2\left[\epsilon + m_p(\alpha) + \frac{\gamma + \gamma N\epsilon}{2N^2\alpha^2} + 3\tau\right]^{1/2}$$

**Proof** We construct a partition $\mathcal{P}$ of $\mathcal{X} \setminus M_p(\alpha)$ with intervals by intersecting the intervals that make up $C_p(\alpha)$ with a partitioning of $\mathcal{X}$ into $N$ intervals of equal probability mass. Let $\{x_1, x_2, \ldots, x_M\}$ be the endpoints of these intervals. There are no more than $N - 1 + 2c_p(\alpha)$ distinct endpoints in $\mathcal{I}$, and we choose $N$ such that $M = N - 1 + 2c_p(\alpha)$.

We bound $(h_1 - h_2)^2$ on each set of the partition $M_p(\alpha) \cup \bigcup_{I \in \mathcal{P}} I$ of $\mathcal{X}$. On $M_p(\alpha)$ $[h_1(1|x) - h_2(1|x)]^2 \leq 1$ trivially. On each $I \in \mathcal{P}$ we must resort to Lemma 3 to derive an upper bound.

Let $I = (u, v)$. Note that for $x \in I$, $[h_1(1|x) - h_2(1|x)]^2 \leq 2[h_1(1|x) - h_1(1|u)]^2 + 2[h_2(1|u) - h_2(1|x)]^2 + 3\tau$. Thus it suffices to bound the variation of each $h$ on $(u, x)$. This is exactly what Lemma 3 provides:

$$[h(1|x) - h(1|u)]^2 \leq \frac{R_u^x(h)}{4}\int_u^x \frac{dx'}{p(x')} \leq \frac{R_u^v(h)}{4}\int_u^x \frac{dx'}{p(x')}$$

where $R_a^b(h)$ is the information regularizer of $h$ on $(a, b)$. Thus $[h_1(1|x) - h_2(1|x)]^2 \leq 3\tau + \frac{1}{2}(R_u^v(h_1) + R_u^v(h_2))\int_u^x dx'/p(x')$. Combining this result with a similar application of Lemma 3 on $(x, v)$ leads to $[h_1(1|x) - h_2(1|x)]^2 \leq 3\tau + (R_u^v(h_1) + R_u^v(h_2))/4 \cdot \int_u^v dx/p(x)$. Since $1/p \leq p/\alpha^2$ on $I$, for $x \in I$ we have

$$[h_1(1|x) - h_2(1|x)]^2 \leq 3\tau + \frac{R_u^v(h_1) + R_u^v(h_2)}{4N\alpha^2} \quad (24)$$

To obtain the bound on $|\mathrm{E}[L_{h_1}] - \mathrm{E}[L_{h_2}]|$ take expectation over $I$ of (24), use $\sum_I R_I(h) < \gamma$, $\int_I p \leq 1/N$, then apply Lemma 4. For the second part of the theorem, we upper bound $\frac{1}{n}\sum(h_1(1|x_i) - h_2(1|x_i))^2$ using (24) in terms of the fraction $f_I$ of samples that fall in interval $I$, and the fraction $f_0$ of samples that fall in $M_\alpha(p)$. Since $\max_I f_I < 1/N + \epsilon$ and $f_0 < m_\alpha(p) + \epsilon$ with probability at least $1 - (M+1)\exp(-2\epsilon^2 n)$, the conclusion follows. □

### Acknowledgments

Thanks to Brendan Frey and Martin Szummer for helpful comments and discussions.

### References

Abe, N., J. Takeuchi, and M. Warmuth (2001). Polynomial learnability of stochastic rules with respect to the KL-divergence and quadratic distance. *IEICE Trans. Inf. & Syst.* E84-D(3): 299–316.

Haussler, D. (1990). Decision theoretic generalizations of the pac learning model. In Arikawa, S., S. Goto, S. Ohsuga, and T. Yokomori, editors, *Algorithmic Learning Theory*, pp. 21–41, New York. Springer Verlag.

Kearns, M. J. and R. E. Schapire (1994). Efficient distribution-free learning of probabilistic concepts. In Hanson, S. J., G. A. Drastal, and R. L. Rivest, editors, *Computational Learning Theory and Natural Learning Systems, Volume I: Constraints and Prospect*, Vol. 1. MIT Press, Bradford.

McDiarmid, C. (1989). On the method of bounded differences. In *Survey in Combinatorics*, pp. 148–188. Cambridge Univ. Press.

Szummer, M. and T. Jaakkola (2003). Information regularization with partially labeled data. In *NIPS'2002*, Vol. 15.